\documentclass[runningheads]{llncs}
\usepackage[T1]{fontenc}
\usepackage{newtxtext}       %
\usepackage[varvw]{newtxmath}       

\usepackage{tabularx}
\usepackage{booktabs} 
\usepackage{caption,subcaption}
\usepackage{pgfplots}
\usepackage{placeins}
\usepackage[hidelinks]{hyperref}

\usepackage{graphicx}
\begin{document}
\title{Incorporating argumentation frameworks for LLM decision-making: Task-specific design and evaluation}
\titlerunning{Argumentation-Augmented LLM Decision-Making}
%
\author{Bochen Han\inst{1,2} \and
Anshu Xiong\inst{1,2} \and
Mingyang Ma\inst{1,2}}
\authorrunning{B. Han et al.}
%
\institute{State Key Laboratory of Mathematical Sciences, Academy of Mathematics and Systems Science, Chinese Academy of Sciences, Beijing 100190, CHINA \and
School of Mathematical Sciences, University of Chinese Academy of Sciences, Beijing 100190, CHINA
\\
\email{hanbochen22@mails.ucas.ac.cn}}
\maketitle              
\begin{abstract}
Argumentation frameworks (AFs) provide explicit, formal mechanisms for representing and resolving conflicts and thus have been incorporated into large language model (LLM) decision-making. The benefits vary and depend on how arguments and their relations are constructed, evaluated, and incorporated into the decision process. {We abstract a workflow consisting of four stages based on existing argumentation-augmented methods} and investigate their design choices over tasks of debate-based decision-making and retrieval-based fact verification. Our evaluation with recent LLMs shows that existing methods do not consistently improve performance across the considered tasks. Motivated by this observation, we develop two task-specific methods. The first is called Qnt-Arg, which organizes arguments around the original speakers in debate materials and aggregates model-estimated argument strengths and candidate-answer priors under gradual semantics. The second is Qlt-Arg that uses credibility estimates to establish preferences between conflicting evidence items and applies grounded semantics to select acceptable evidence for subsequent LLM inference. We examine the roles of generative and decision models within the two methods. In the debate setting, Qnt-Arg consistently outperforms other argumentation-augmented methods across matched configurations; Qlt-Arg improves retrieval-based fact-verification performance compared with the baselines. These findings demonstrate the value of combining formal argumentation and LLM while highlighting the importance of aligning with tasks in argumentation construction, model selection, and the role of formal semantic reasoning.

\keywords{Argumentation Frameworks  \and Large Language Models \and Decision-making}
\end{abstract}
\section{Introduction}
Large language models (LLMs) increasingly support decision-making tasks that require weighing competing reasons and reconciling conflicting information. These reasons may come from participants in a discussion, interacting agents, or retrieved sources. Multi-agent debate elicits alternative responses and reasoning processes \cite{du2024debate}, while retrieval-based fact verification draws on evidence whose credibility may vary \cite{ge2025confact}. Access to multiple perspectives creates a need to determine how each contribution should affect the decision. Studies of knowledge conflicts show that LLM judgments can be influenced by confirmation bias when external evidence conflicts with their existing knowledge \cite{xie2024conflicts}. This motivates explicit representations of argumentative relationships and formal procedures for evaluating them.

Formal argumentation represents arguments and their relations and specifies semantics for determining acceptability or strength \cite{dung1995acceptability,atkinson2017argumentation}. Recent argumentation-augmented methods connect these formal methods to LLMs. Quantitative methods such as ArgLLM generates supporting and attacking arguments and evaluates a target claim using gradual semantics \cite{freedman2025argllm}. Moreover, ArgRAG tasks advantages of retrieved evidence in constructing a quantitative bipolar argumentation framework (QBAF) from retrieved evidence and computes the claim's final strength \cite{zhu2025argrag}. Qualitative methods such as ArgCheck incorporates abstract argumentation framework (AF) and its grounded semantics to identify accepted evidence for a subsequent model judgment \cite{chen2026argcheck}. The LLM decision-making methods give formal reasoning two distinct roles: computing a candidate answer's strength and determining the acceptance of information used by a model.

The effectiveness of these roles depends on the connection between natural language inputs and formal semantic reasoning. Debate transcripts contain identifiable speakers, their arguments, and the relations between their arguments~\cite{tan2016winning,liang2024debatrix,lan2026cedar,zhang2016iq2,chang2020convokit,mayfield2019afd}. Retrieved materials provide evidence items whose credibility and compatibility must be assessed~\cite{ge2025confact,schlichtkrull2023averitec,huang2026deepfact}. Our evaluation in Section~\ref{sec:qnt-results} and ~\ref{sec:qlt-results} finds that ArgLLM underperforms than direct LLM reasoning on debate datasets and ArgRAG performs worse than direct retrieval-augmented LLM generation on some fact-verification datasets. This motivates us to conduct an examination of existing works that incorporate formal argumentation into LLM decision-making. Based on their commonalities, we present a general workflow that consists of four main stages: argumentation construction, acceptability assessment, formal semantic reasoning under argumentation semantics, and decision-making. The differences lie in the used argumentative formalism and its semantics and roles that LLM takes across the four stages. By such a general workflow, we can better study the impact of various design choices targeting specific tasks.  

Based on the general workflow, we develop two argumentation-augmented LLM methods around these choices. For debate-based decision-making, \emph{Qnt-Arg} proposes speaker-conditioned argumentation construction: agents generate reasons from the perspectives and statements of the speakers explicitly involved in the debate documents. Gradual semantics is used to combine argument strengths, relation weights, and a candidate-answer prior to determine the final prediction. For retrieval-based fact verification, \emph{Qlt-Arg} treats evidence items as arguments and uses credibility estimates to orient attacks between conflicting items. Grounded semantics is used to compute grounded acceptance annotations as auxiliary decision-making information. We use two types of LLMs, ordinary LLMs that generate natural language texts as output and decision models that return scores or probabilities for predefined answer options.

Across six debate datasets, Qnt-Arg achieves 78.41 overall accuracy, compared with 74.38\% by direct LLM inference and 76.99\% by the strongest direct baseline. On retrieval-based tasks, Qlt-Arg with grounded-extension (GE) filtering outperforms RAG LLM in all six Top-5/Top-10 settings, reaching 80.50\% on AVeriTeC (Top-10) and 87.50\% on DeepFact-Bench (Top-5). Applying Qlt-Arg without filtering remains more effective on CONFACT. Results across datasets and LLM assignments show the effectiveness of task-specific configurations.

Our main contributions are as follows:
\begin{enumerate}
\item We propose a general workflow of argumentation-augmented LLM and develop two methods Qnt-Arg and Qlt-Arg based on it. Qnt-Arg incorporates quantitative AF to LLM decision-making tasks with multi-speaker debate content, and combines speaker-conditioned argumentation construction with quantitative reasoning under gradual semantics. Qlt-Arg is designed for retrieval-based
decision-making, which combines credibility-based preferences with grounded semantics of qualitative AF to identify acceptable evidence and guide model inference.
\item We evaluate two methods on both debate-based decision-making and retrieval-based fact verification tasks, comparing local assessments, candidate-answer priors, and semantic outputs. The results show that Qnt-Arg and Qlt-Arg respectively excel in debate datasets and retrieval-based datasets when compared with direct LLM and decision model, ArgLLM, ArgRAG, and other relevant baselines. Nevertheless, no single method surpasses across all tasks, illustrating the necessity of making task-specific design choices within various stages of combining formal argumentation with LLM decision-making.
\end{enumerate}

\section{Related Work}
\label{sec:related-work}

\paragraph{Quantitative argumentation and LLMs.}
ArgLLM combines LLM-generated argument trees with model-estimated base strengths and gradual semantics \cite{freedman2025argllm}. Its construction requests supporting and attacking arguments for each target. Qnt-Arg instead conditions construction on the speakers already present in debate materials and examines how local assessments interact with candidate-answer priors. Zhou et al. \cite{zhou-etal-2025-evaluating} study uncertainty quantification in ArgLLM; our model-role comparisons examine how generative and decision models contribute at different assessment stages.

ArgRAG constructs QBAFs from retrieved passages, annotates claim--evidence and evidence--evidence relations, and evaluates claim strengths under QE semantics \cite{zhu2025argrag}. Yin et al. \cite{yin2025contestability} study contestability in EW-QBAFs through adjustments to relation weights. Their weighted semantics provides the formal basis for separating argument assessments from relation assessments in our quantitative configurations.

ARGORA combines expert-generated QBAFs with judge-aligned overrides \cite{jin2026argora}, while ArgEval instantiates reusable QBAFs from policy documents \cite{dejl2026argeval}.

\paragraph{Qualitative argumentation for evidence assessment.}
Grounded semantics gives a unique, conflict-free extension for an AF \cite{dung1995acceptability}. ArgCheck combines evidence collection, source-priority relations, and grounded semantics to guide fact-checking \cite{chen2026argcheck}. LLM-ASPIC$^{+}$ uses extracted rule preferences and a grounded extension to guide the final LLM \cite{fang2025llmaspic}. Qlt-Arg uses numerical credibility estimates to orient conflicts and supplies evidence acceptance annotations to the final LLM. Our experiments compare this use of qualitative acceptance with quantitative prediction and with final-model inputs containing credibility scores or acceptance annotations separately. This places the role of formal semantic reasoning at the center of the comparison.

\section{Preliminaries}
\label{sec:preliminaries}

We introduce the argumentation frameworks and semantics used in our methods. Throughout, $\mathcal{A}$ denotes a finite set of arguments.

\subsection{Abstract Argumentation Frameworks and Grounded Semantics}
\label{sec:af}

An abstract argumentation framework (AF) is a pair $\mathcal{F}=(\mathcal{A},\mathcal{R}^{-})$, where $\mathcal{R}^{-}\subseteq\mathcal{A}\times\mathcal{A}$ is a directed attack relation \cite{dung1995acceptability}. We write $\operatorname{Att}(a)=\{b\in\mathcal{A}:(b,a)\in\mathcal{R}^{-}\}$. A set $S\subseteq\mathcal{A}$ is conflict-free if $(S\times S)\cap\mathcal{R}^{-}=\emptyset$, and $S$ defends $a$ if for each attacker $b$ of $a$ there exists an argument in $S$ that attacks $b$. The characteristic function collects the arguments defended by $S$:
\begin{equation}
\Gamma_{\mathcal{F}}(S)=
\{a\in\mathcal{A}:\forall b\in\operatorname{Att}(a),\
\operatorname{Att}(b)\cap S\neq\emptyset\}.
\label{eq:characteristic-function}
\end{equation}
The grounded extension $E_{\mathrm{gr}}(\mathcal{F})$ is the least fixed point of $\Gamma_{\mathcal{F}}$ under set inclusion. For finite AFs, it is obtained by iterating
\begin{equation}
E_0=\emptyset,\qquad E_{k+1}=\Gamma_{\mathcal{F}}(E_k)
\label{eq:grounded-iteration}
\end{equation}
until $E_{k+1}=E_k$. This unique, conflict-free extension contains the arguments accepted under grounded semantics.

\subsection{Quantitative Bipolar Argumentation Frameworks}
\label{sec:qbaf}
\label{sec:ew-qbaf}

A quantitative bipolar argumentation framework (QBAF) is a quadruple $\mathcal{Q}=(\mathcal{A},\mathcal{R}^{-},\mathcal{R}^{+},\beta)$, with disjoint attack and support relations $\mathcal{R}^{-},\mathcal{R}^{+}\subseteq\mathcal{A}\times\mathcal{A}$ and base strength function $\beta:\mathcal{A}\rightarrow[0,1]$. An edge-weighted QBAF (EW-QBAF) additionally assigns weights to relations $w:\mathcal{R}^{-}\cup\mathcal{R}^{+}\rightarrow[0,1]$ \cite{yin2025contestability}; setting $w\equiv1$ recovers the unweighted case. Let $\operatorname{Sup}(a)=\{b:(b,a)\in\mathcal{R}^{+}\}$.

For some strength assignment $s:\mathcal{A}\rightarrow[0,1]$, write $v_{ba}=w(b,a)s(b)$. The DF-QuAD semantics and quadratic energy (QE) semantics aggregate incoming contributions as
\begin{equation}
\begin{aligned}
g_{\mathrm{DF\text{-}QuAD}}^{w}(a;s)
&=\prod_{b\in\operatorname{Att}(a)}(1-v_{ba})
-\prod_{b\in\operatorname{Sup}(a)}(1-v_{ba}),\\
g_{\mathrm{QE}}^{w}(a;s)
&=\sum_{b\in\operatorname{Sup}(a)}v_{ba}
-\sum_{b\in\operatorname{Att}(a)}v_{ba}.
\end{aligned}
\label{eq:weighted-aggregation}
\end{equation}
With $z_+=\max\{0,z\}$, the influence functions of the two types of semantics are $h_{\mathrm{DF\text{-}QuAD}}(z)=z_+$ and $h_{\mathrm{QE}}(z)=z_+^2/(1+z_+^2)$ \cite{rago2016discontinuity,potyka2018continuous}. Both use the update operator
\begin{equation}
\begin{aligned}
\bigl[T_m^{w}(s)\bigr]_a
=\beta(a)-\beta(a)h_m(-g_m^{w}(a;s))+(1-\beta(a))h_m(g_m^{w}(a;s)),
\end{aligned}
\label{eq:unified-update}
\end{equation}
where $m\in\{\mathrm{DF\text{-}QuAD},\mathrm{QE}\}$ and $\bigl[T_m^{w}(s)\bigr]_a$ is the updated strength of $a$, with $\beta$ fixed. Empty products equal $1$ and empty sums equal $0$, so arguments without incoming relations retain their base strengths.

Acyclic graphs compute final strengths $\sigma$ in topological order. For graphs that may contain cycles, continuous QE uses \cite{potyka2018continuous,zhu2025argrag}
\begin{equation}
\frac{d s_a(t)}{dt}=[T_{\mathrm{QE}}^{w}(s(t))]_a-s_a(t),
\qquad s_a(0)=\beta(a).
\label{eq:continuous-qe}
\end{equation}
At convergence, final strengths $\sigma=\lim_{t\rightarrow\infty}s(t)$ satisfy $\sigma=T_{\mathrm{QE}}^{w}(\sigma)$.

\section{Method}
\label{sec:method}

\subsection{Task Formulation and the General Workflow}
\label{sec:overall-design}

An input $x=(q,C)$ consists of a question or claim $q$ and its task context $C$, such as a debate transcript. Retrieved evidence, when used, is denoted separately by $\mathcal{E}$. We formulate candidate outcomes as binary claims, with predictions $\hat{y}\in\{\mathrm{True},\mathrm{False}\}$. 

We abstract a four-stage workflow from existing argumentation-augmented  LLM decision-making methods.
Stage 1: \textit{Argumentation construction} forms arguments and determines their relations.
Stage 2: \textit{Acceptability assessment} supplies strengths, relation weights, or preferences using LLM outputs, fixed initialization, or explicit priorities.
Stage 3: \textit{Formal semantic reasoning} computes final strengths or acceptance under an argumentation semantics.
Stage 4: \textit{Decision-making} maps semantic outputs to predictions or recommendations through explicit decision rules or an LLM judgment. 

\begin{figure}[t]
\centering
\includegraphics[width=\linewidth]{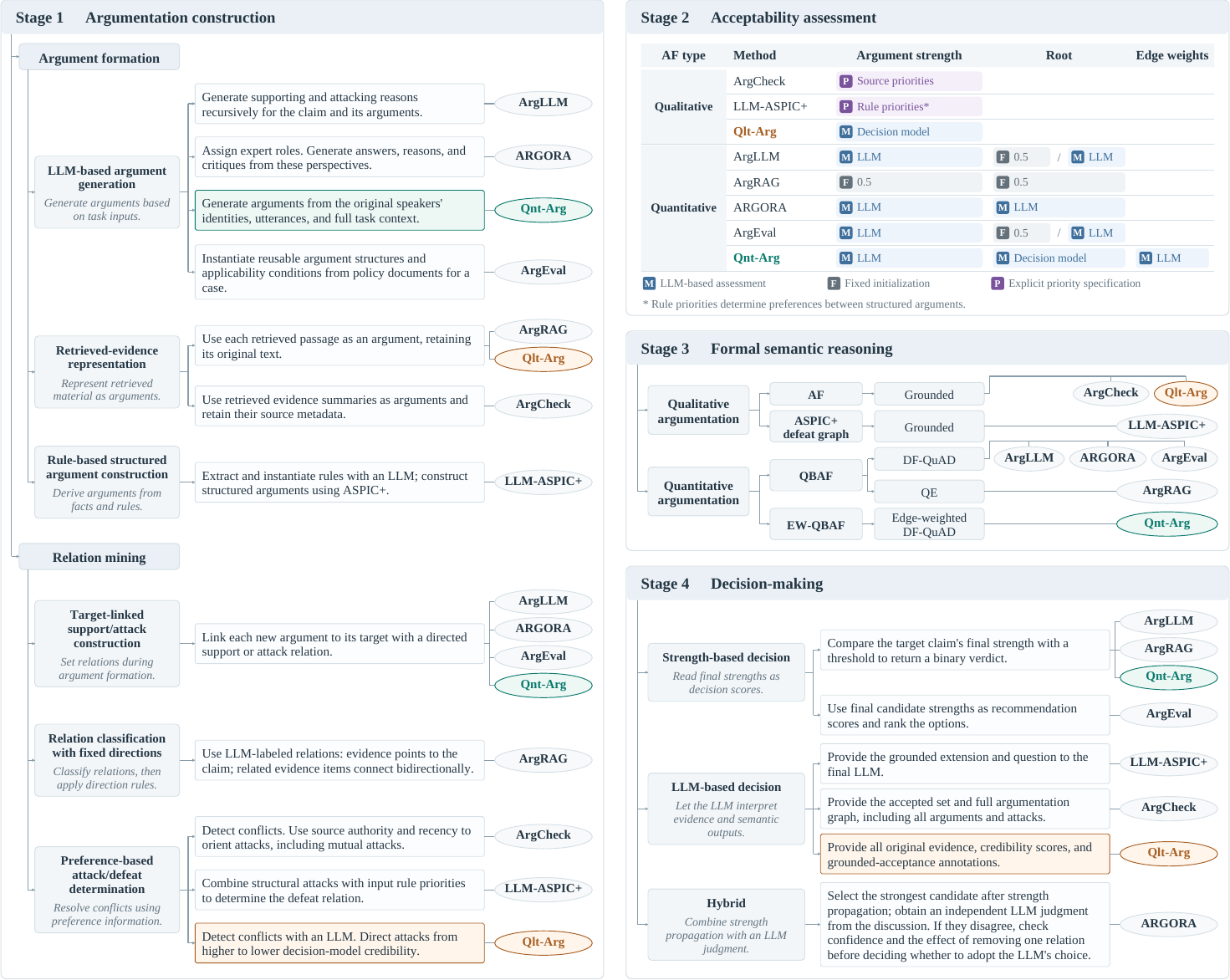}
\caption{The four-stage workflow and the design choices of various methods.}
\label{fig:workflow}
\end{figure}

\begin{figure}[t]
\centering
\includegraphics[width=\linewidth]{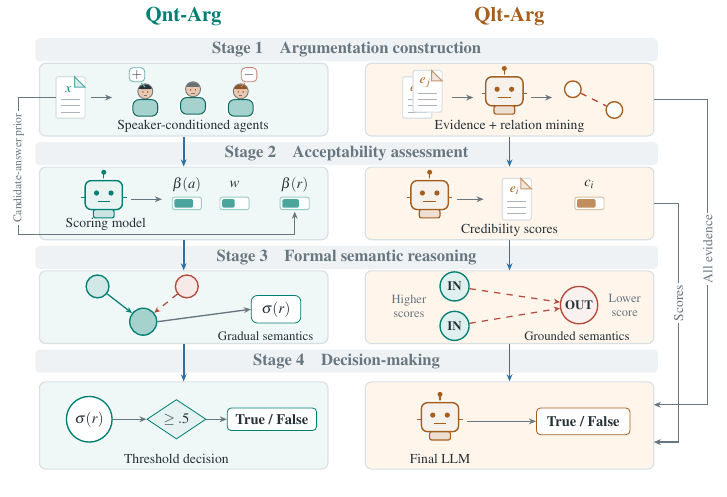}
\caption{The main Qnt-Arg and Qlt-Arg instantiations of the general workflow.}
\label{fig:method-overview}
\end{figure}

In Figure~\ref{fig:workflow}, existing methods ~\cite{freedman2025argllm,zhu2025argrag,jin2026argora,dejl2026argeval,chen2026argcheck,fang2025llmaspic} and the two methods proposed in this paper are shown according to their design choices in the workflow. Qnt-Arg features generation conditioned on speakers in debate datasets, separate argument and relation assessments, and candidate-answer priors given by a decision model. Qlt-Arg uses credibility scores by a decision model to orient conflicts and supplies all evidence, credibility scores, and grounded acceptance annotations to the final LLM. Figure~\ref{fig:method-overview} illustrates processes of the two methods.

\subsection{Qnt-Arg: Speaker-Conditioned Quantitative Argumentation}
\label{sec:qt-arg}

Qnt-Arg has novelty in argumentation construction and assessment, whereas it uses the same gradual semantics as ArgLLM\cite{freedman2025argllm} to compute quantitative reasoning results and uses the implementation developed in \cite{zhou-etal-2025-evaluating} for uncertainty quantification. In the debate tasks claim $q$ specifies a candidate outcome, while $C$ contains the transcript, judgment criterion, and relevant background or target information. The root argument $r$ represents $q$.

\paragraph{Speaker-conditioned argumentation construction.}
Let $u_1,\ldots,u_N$ be the selected speakers. Each agent uses the same generative LLM, conditioned on the complete task input, a speaker identity, and that speaker's utterances in chronological order. Agents generate arguments from these perspectives and may develop further reasons using model knowledge.

At depth one ($D=1$), agent $i$ generates one self-contained argument $a_i$ and declares whether it supports or attacks the root $r$. We add the corresponding edge $(a_i,r)$ to $\mathcal{R}^{+}$ or $\mathcal{R}^{-}$. This yields the first-level argument set $\mathcal{A}_1=\{a_1,\ldots,a_N\}$.

At depth two ($D=2$), each agent makes one supporting call and
one attacking call over the fixed set $\mathcal{A}_1$. For agent $i$, these calls generate distinct arguments $b_i^{+}$ and $b_i^{-}$ and select their respective target indices $j_i^{+},j_i^{-}\in\{1,\ldots,N\}$. We add the edges
$
(b_i^{+},a_{j_i^{+}})\in\mathcal{R}^{+},
(b_i^{-},a_{j_i^{-}})\in\mathcal{R}^{-}.
$
For greater depths, the same procedure is repeated over the fixed
set of arguments generated at the preceding level.
Targets may include the agent's own upper-level argument. Both calls use the same upper-level set. The resulting graph is acyclic, with one parent per non-root argument. Construction produces $1+(2D-1)N$ nodes at depth $D$.

\paragraph{Assessment and computation under gradual semantics.}
In the unweighted configuration, each non-root base strength jointly assesses the argument's contextual validity and the validity of its relation to its parent. The edge-weighted configuration separates these assessments: argument validity determines the base strength, and relation validity determines the edge weight.

The candidate-answer prior initializes the root:
\begin{equation}
\beta(r)=
\begin{cases}
0.5, & \text{neutral initialization},\\
d_{M_0}(x), & \text{model-based initialization}.
\end{cases}
\label{eq:qt-root-prior}
\end{equation}
Here, $d_{M_0}(x)$ is the score assigned to the candidate answer by LLM $M_0$ according to the original task input. The prior model and local assessment model can be chosen independently.

The DF-QuAD or edge-weighted DF-QuAD semantics evaluates the constructed argumentation graph in topological order. At depth $D$, computation proceeds from depth $D$ through the preceding levels to the root. The result is $\mathrm{True}$ when $\sigma(r)\geq0.5$ and $\mathrm{False}$ otherwise.

\subsection{Qlt-Arg: Grounded Acceptance for Retrieved Evidence under Qualitative Argumentation}
\label{sec:ql-arg}

QltArg features credibility-based determination of attack directions between evidence items when constructing relations. After argumentation construction Qlt-Arg shares the similar formal semantic reasoning and decision-making processes with ArgCheck\cite{chen2026argcheck}.

\paragraph{Evidence and relation construction.}
The task input $x=(q,C)$ contains the question or claim and its context. Retrieved evidence forms an ordered list $\mathcal{E}=(e_1,\ldots,e_m)$. Each evidence item becomes one argument. The main configuration retains all retrieved items in their original form and order.

A relation model jointly classifies evidence pairs as supporting, contradicting, or neither. Support and contradiction labels are mutually exclusive; uncertain or mixed pairs are omitted. Let $\mathcal{CT}$ denote the set of unordered contradiction pairs.

\paragraph{Credibility preferences and acceptance under grounded semantics.}
Each evidence argument $e_i$ is assigned a credibility score $c_i$. A contradiction is oriented from the more credible item to the less credible one, yielding the AF $\mathcal{F}=(\mathcal{A}_E,\mathcal{R}^{-})$, where $\mathcal{A}_E=\{e_1,\ldots,e_m\}$ and
\begin{equation}
\mathcal{R}^{-}
=\{(e_i,e_j):\{e_i,e_j\}\in\mathcal{CT},\ c_i>c_j\}.
\label{eq:credibility-attacks}
\end{equation}
Ties induce no attack. Support relations and candidate-answer roots are excluded from this evidence AF. Credibility strictly decreases along each attack, so the graph is acyclic. We compute its grounded extension $E_{\mathrm{gr}}(\mathcal{F})$ as defined in Section~\ref{sec:af}.

The final LLM receives the task input, all retrieved evidence, their credibility scores, and the identifiers of evidence items in $E_{\mathrm{gr}}(\mathcal{F})$, and gives the final answer. Grounded acceptance thus provides relational information alongside the evidence used for the final judgment.

\section{Experiments}

\subsection{Experimental Setup}
\label{sec:experimental-setup}

\textbf{Tasks and datasets.}
For debate-based decision-making tasks, each input contains the full available debate or discussion, its judgment criterion, and a candidate outcome proposition. CMV~\cite{tan2016winning} tests whether a designated reply branch changes the original poster's view, with one instance per branch. DebateArt~\cite{liang2024debatrix} tests whether Pro wins (original labels $0/1$ map to $\mathrm{True}/\mathrm{False}$); CEDAR~\cite{lan2026cedar} whether the affirmative side wins; IQ2~\cite{zhang2016iq2} whether the \textit{For} side achieves the larger audience-support increase; and Supreme Court~\cite{danescu2012echoes,chang2020convokit} whether the petitioner wins.
Moreover, Wikipedia AfD~\cite{mayfield2019afd} uses 250 discussions. Each produces one proposition for its true category and one for a randomly selected incorrect category among keep, delete, merge, and redirect, yielding 500 instances. Explicit outcome disclosures, including delta awards, winner announcements, and voting results, are removed from all inputs.

For retrieval-based fact verification tasks, \textit{CONFACT}~\cite{ge2025confact} provides webpages and \textit{AVeriTeC}~\cite{schlichtkrull2023averitec} provides candidate sentences, which are processed and ranked for each instance. \textit{DeepFact-Bench}~\cite{huang2026deepfact} uses web search through Serper/Google, followed by content retrieval and passage ranking.

\textbf{Models and configurations.}
The main experiments use DeepSeek-V4-Flash as the generative model. Within each configuration, the same generative model performs argument generation, relation annotation, and final answering wherever the model applies. Perplexity Decider and Jev serve as decision models in the debate and retrieval experiments, respectively. They provide scores for local assessments, candidate-answer priors, and direct decision baselines.

We compare neutral candidate-answer priors with direct generative-model and decision-model priors. Prior comparisons reuse the same generated arguments, relations, and non-root assessments. For Qlt-Arg, we compare final-model inputs containing credibility scores, grounded acceptance annotations, or both, using the same retrieved evidence.

\textbf{Baselines and alternative decision procedures.}
\emph{Direct LLM} and \emph{Direct Decision Model} evaluate the original task input without additional arguments or retrieved evidence. Experiments for debate-based tasks compare Qnt-Arg with \emph{ArgLLM} \cite{freedman2025argllm}, implemented using the code of Zhou et al. \cite{zhou-etal-2025-evaluating}, as well as \emph{CoT} and five-sample \emph{Majority Voting}. Over the generated speaker-conditioned arguments, \emph{Speaker Voting} aggregates first-level stances. \emph{Text-Judge} provides the argument texts to an LLM, while \emph{Graph-Judge} additionally provides relations, local assessments, and the candidate-answer prior.

For retrieval tasks, \emph{RAG LLM} and \emph{RAG Decision Model} answer directly from retrieved evidence. \emph{ArgRAG} \cite{zhu2025argrag} uses uniform base strengths of $0.5$; its adaptations add model-estimated base strengths and relation weights. All methods within a retrieval setting start from the same retrieved evidence list.
We also examine cross-task instantiations: Qnt-Arg evaluating retrieved evidence under edge-weighted QE semantics, and Qlt-Arg providing credibility scores and grounded acceptance annotations over cached speaker-conditioned arguments.

\textbf{Implementation details.}
DeepSeek-V4-Flash is accessed through DeepSeek's official API,\footnote{\url{https://api-docs.deepseek.com/}} Jev through TypeSafe's official API,\footnote{\url{https://api.typesafe.ai/redoc}} and Perplexity Decider through Perplexity's official Decisions API.\footnote{\url{https://docs.perplexity.ai/docs/decisions/quickstart}} The generative model uses non-thinking mode for argument generation, relation annotation, and final answering. Quantitative reasoning uses Uncertainpy argumentation library\footnote{\url{https://github.com/nicopotyka/Uncertainpy}}and its edge-weighted extensions. Continuous QE reasoning uses RK4 with step size $0.1$ and a maximum-derivative stopping threshold of $0.001$. Binary score-based results are positive when the final score is at least $0.5$. 

\subsection{Qnt-Arg Results and Analysis}
\label{sec:qnt-results}

\begin{table}[!t]
\centering
\caption{Accuracy (\%) comparison of all baselines and Qnt-Arg on six debate datasets. Overall accuracy pools all 2,186 binary instances from the six datasets.}
\label{tab:qtarg-main-results}
\captionsetup[subtable]{position=top,skip=3pt}
\begin{subtable}{\linewidth}
\centering
\caption{Baseline and Qnt-Arg results.}
\label{tab:qnt-dataset-results}
\small
\setlength{\tabcolsep}{3pt}
\renewcommand{\arraystretch}{1.08}
\begin{tabular*}{\linewidth}{@{\extracolsep{\fill}}lccccccc@{}}
\toprule
Method & CMV & DA & CEDAR & IQ2 & SC & AfD & Overall \\
\midrule
Direct LLM & 68.47 & 70.21 & 58.28 & 59.05 & 77.00 & 97.40 & 74.38 \\
Direct Decision Model & 70.28 & \textbf{77.66} & 66.46 & \textbf{60.00} & 76.60 & \textbf{97.80} & 76.99 \\
CoT & 69.28 & 70.21 & 58.08 & 59.05 & 81.60 & 97.60 & 75.62 \\
Majority Voting & 69.08 & 74.47 & 60.12 & 56.19 & \textbf{83.60} & 97.20 & 76.44 \\
\midrule
\multicolumn{8}{@{}l}{\textit{Depth $D=1$}} \\
ArgLLM & 62.05 & 61.70 & 51.74 & 53.33 & 64.60 & 87.40 & 65.69 \\
Qnt-Arg & \textbf{72.29} & \textbf{77.66} & 66.87 & 59.05 & 80.20 & \textbf{97.80} & 78.32 \\
\midrule
\multicolumn{8}{@{}l}{\textit{Depth $D=2$}} \\
ArgLLM & 61.24 & 63.83 & 51.74 & 48.57 & 65.20 & 83.60 & 64.64 \\
Qnt-Arg & 70.88 & 76.60 & \textbf{67.89} & \textbf{60.00} & 81.00 & \textbf{97.80} & \textbf{78.41} \\
\bottomrule
\end{tabular*}
\end{subtable}
\par\smallskip
\begin{subtable}{\linewidth}
\centering
\caption{Overall accuracy of alternative decision procedures.}
\label{tab:qtarg-decision-procedure}
\small
\renewcommand{\arraystretch}{1.08}
\begin{tabular*}{\linewidth}{@{\extracolsep{\fill}}lrrrr@{}}
\toprule
Depth & Speaker Voting & Text-Judge & Graph-Judge & Qlt-Arg \\
\midrule
$D=1$ & 71.82 & 75.07 & 74.84 & 75.48 \\
$D=2$ & 71.82 & 73.92 & 75.07 & 75.39 \\
\bottomrule
\end{tabular*}
\end{subtable}
\par\smallskip
\begin{minipage}{\linewidth}
\footnotesize
DA: DebateArt; SC: Supreme Court; AfD: Wikipedia AfD.
ArgLLM uses joint argument--relation assessments and a neutral candidate-answer prior of $0.5$.
Qnt-Arg uses speaker-conditioned argument graphs, DeepSeek-V4-Flash local assessments, edge-weighted DF-QuAD, and a Perplexity Decider candidate-answer prior.
Bold denotes the highest value in each column, including ties.
\end{minipage}
\end{table}

\textbf{Main results.}
Table~\ref{tab:qnt-dataset-results} reports the results on debate-based datasets. Qnt-Arg achieves 78.32\% overall accuracy at $D=1$ and 78.41\% at $D=2$. The latter exceeds Direct LLM by 4.03 pp and Direct Decision Model by 1.42 pp, and also exceeds CoT and Majority Voting. The evaluated ArgLLM configuration obtains 65.69\% and 64.64\% at the two depths, respectively. These results demonstrate the effectiveness of the complete Qnt-Arg configuration on the debate-based decision-making tasks.

The effect of depth varies by dataset. Accuracy increases on CEDAR, IQ2, and Supreme Court, decreases on CMV and DebateArt, and remains unchanged on Wikipedia AfD. Overall moving from D=1 to D=2 improves 0.09 pp in accuracy. Majority Voting is strongest on Supreme Court, with 83.60\% accuracy compared with 81.00\% for Qnt-Arg at $D=2$. 

\begin{table}[!tbp]
\centering
\caption{Qnt-Arg ablations on debate tasks at $D=2$. $\Delta$ is the reported change in percentage points(pp) from the complete configuration.}
\label{tab:qtarg-ablation}
\small
\renewcommand{\arraystretch}{1.08}
\begin{tabular*}{\linewidth}{@{\extracolsep{\fill}}ccrr@{}}
\toprule
Candidate-answer prior & \shortstack{Separate argument and\\relation assessments} & Overall (\%) & $\Delta$ (pp) \\
\midrule
$\checkmark$ & $\checkmark$ & \textbf{78.41} & $0.00$ \\
--- & $\checkmark$ & 74.75 & $-3.66$ \\
$\checkmark$ & --- & 76.72 & $-1.69$ \\
--- & --- & 72.37 & $-6.04$ \\
\bottomrule
\end{tabular*}
\end{table}

\textbf{Ablation study.}
Table~\ref{tab:qtarg-ablation} examines Qnt-Arg at $D=2$. Replacing the candidate-answer prior with a neutral root strength of $0.5$ reduces overall accuracy from 78.41\% to 74.75\%. Replacing separate argument and relation assessments with joint base-strength assessments and unit relation weights reduces accuracy from 78.41\% to 76.72\%. Applying both replacements yields the lowest accuracy at 72.37\%. These show the necessity of the design of Qnt-Arg in candidate answer prior and separate assessments for arguments and relations.

\textbf{Model roles.}
With a generative model for local assessments, switching the candidate-answer prior from the generative model to the decision model raises accuracy from 75.30\% to 78.41\%. Using the decision model for all acceptability assessments yields 76.53\%, while averaging the two direct prediction scores yields an accuracy below 78.41\%. These comparisons favor generative models for local assessment and decision models for candidate-answer priors.

\textbf{Executing formal semantic reasoning.}
The alternative decision procedures in Table~\ref{tab:qtarg-decision-procedure} use the same cached arguments as Qnt-Arg. At $D=2$, Qnt-Arg achieves 78.41\%, compared with 73.92\% for Text-Judge and 75.07\% for Graph-Judge. Graph-Judge receives the argument graph, local assessments, relation weights, and the same decision model candidate-answer prior. Qlt-Arg reduces accuracy from 78.41\% to 75.39\% using credibility scores and grounded acceptance annotations. These results indicate that quantitative semantics can be more suitable for debate-based tasks.

\begin{figure}[!tbp]
\centering
\includegraphics[width=\linewidth]{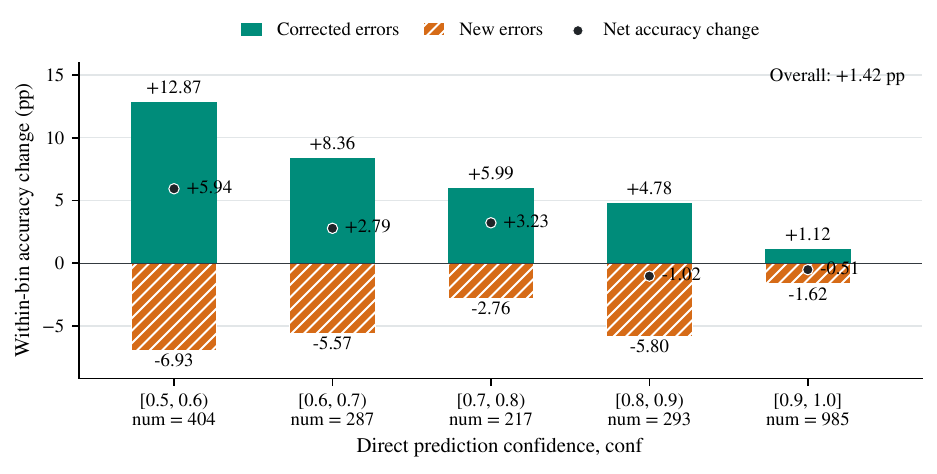}
\caption{Accuracy changes from Direct Decision Model (Perplexity Decider) to Qnt-Arg at $D=2$, grouped by initial confidence $\mathrm{conf}=\max(\beta(r),1-\beta(r))$. Bars show corrections and new errors as percentages of each bin; dots show their net difference.}
\label{fig:qtarg-direct-confidence}
\end{figure}

\textbf{Changes at different initial confidence levels.}
Figure~\ref{fig:qtarg-direct-confidence} examines how reasoning under gradual semantics changes the direct decision-model predictions, showing that Qnt-Arg gains the most when the direct decision model is uncertain (confidence near $0.5$). For $\mathrm{conf}\geq0.8$, Qnt-Arg corrects 25 errors and introduces 33, yielding a decrease of 0.63 pp in accuracy within this group.

\FloatBarrier
\subsection{Qlt-Arg Results and Analysis}
\label{sec:qlt-results}

\begin{table}[!tbp]
\centering
\caption{Accuracy (\%) on CONFACT, AVeriTeC, and DeepFact-Bench with Top-5 and Top-10 retrieved evidence.}
\label{tab:confact-main-results}
\footnotesize
\setlength{\tabcolsep}{3pt}
\renewcommand{\arraystretch}{1.08}
\begin{tabular*}{\linewidth}{@{\extracolsep{\fill}}lrrrrrr@{}}
\toprule
Method & \multicolumn{2}{c}{CONFACT} & \multicolumn{2}{c}{AVeriTeC} & \multicolumn{2}{c}{DeepFact-Bench} \\
\cmidrule(lr){2-3}\cmidrule(lr){4-5}\cmidrule(l){6-7}
& Top-5 & Top-10 & Top-5 & Top-10 & Top-5 & Top-10 \\
\midrule
Direct LLM & 78.56 & 78.56 & 68.00 & 68.00 & 84.00 & 84.00 \\
Direct Decision Model & 65.30 & 65.30 & 72.00 & 72.00 & 83.00 & 83.00 \\
RAG LLM & 76.92 & 78.23 & 76.50 & 78.00 & 84.42 & 83.00 \\
RAG Decision Model & 75.94 & 76.27 & \textbf{79.50} & 80.00 & 86.50 & \textbf{86.00} \\
ArgRAG & 68.74 & 68.69 & 63.50 & 62.31 & 81.00 & 79.50 \\
ArgRAG with model scores & 69.07 & 69.02 & 68.00 & 65.83 & 82.50 & 80.50 \\
Edge-weighted ArgRAG & 68.74 & 69.18 & 69.00 & 68.84 & 83.00 & 81.00 \\
Qnt-Arg & 70.38 & 69.51 & 75.00 & 76.38 & 86.00 & 84.00 \\
\midrule
Qlt-Arg (w/o GE filtering) & \textbf{79.05} & \textbf{80.33} & 75.88 & 77.89 & 81.00 & 82.00 \\
Qlt-Arg (w/ GE filtering) & 77.09 & 78.40 & 78.00 & \textbf{80.50} & \textbf{87.50} & 84.00 \\
\bottomrule
\end{tabular*}
\par\smallskip
\begin{minipage}{\linewidth}
\footnotesize
LLM components use DeepSeek-V4-Flash; decision-model components use Jev. Qnt-Arg uses edge-weighted QE semantics.
GE filtering uses Qlt-Arg when the grounded extension excludes evidence and falls back to RAG LLM otherwise.
Bold denotes the highest accuracy in each column.
\end{minipage}
\end{table}

\textbf{Main results.}
Table~\ref{tab:confact-main-results} reports results on the three retrieval-based datasets. In the Top-10 grounded acceptance analysis, the grounded extension contains all evidence arguments in 73.11\% of CONFACT instances, 81.91\% of AVeriTeC instances, and 93.50\% of DeepFact-Bench instances. In these instances, the grounded acceptance annotations therefore do not differentiate evidence arguments at all. The within-CONFACT analysis further shows a larger benefit when conflicting evidence receives different acceptance statuses: Qlt-Arg improves over RAG Decision Model by 8.54 pp when the grounded extension does not contain all the evidence arguments, compared with 2.47 pp when it contains all arguments. These observations motivate \emph{GE filtering}, which routes each instance to Qlt-Arg when the grounded extension excludes at least one argument and to RAG LLM otherwise. On the Qlt-Arg branch, the final LLM receives all retrieved evidence, credibility scores, and grounded acceptance annotations. The configuration without GE filtering applies Qlt-Arg to all instances.

With GE filtering, Qlt-Arg improves over RAG LLM in all six settings. The gains for Top-5 and Top-10 evidence are 0.17 and 0.17 pp on CONFACT, 1.50 and 2.50 pp on AVeriTeC, and 3.08 and 1.00 pp on DeepFact-Bench. Qlt-Arg achieves the highest accuracy on AVeriTeC with Top-10 evidence (80.50\%) and DeepFact-Bench with Top-5 evidence (87.50\%), exceeding RAG Decision Model by 0.50 and 1.00 pp, respectively.

Compared to applying Qlt-Arg to all instances, GE filtering improves both retrieval settings on AVeriTeC and DeepFact-Bench, with the largest increase of 6.50 pp on DeepFact-Bench with Top-5 evidence. CONFACT instead favors applying Qlt-Arg throughout: the accuracy is 79.05\% and 80.33\% without GE filtering, compared with 77.09\% and 78.40\% with filtering. These results demonstrate the effectiveness of Qlt-Arg and also indicate that designing effective configurations shall take into account of tasks per se.
\begin{figure}[!tbp]
\centering
\includegraphics[width=\linewidth]{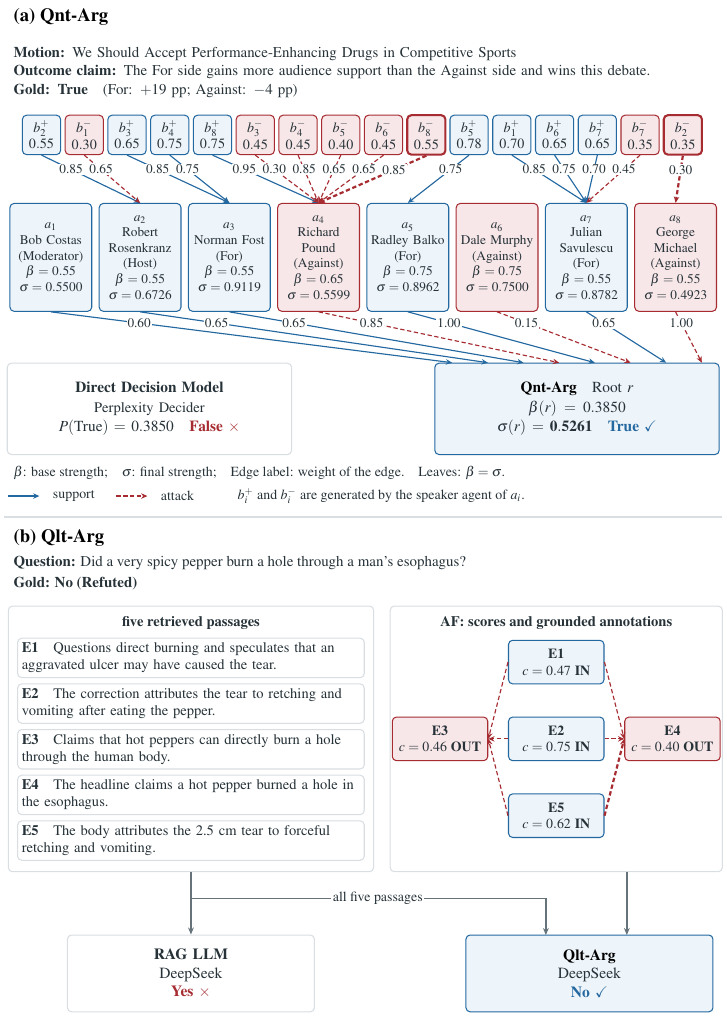}
\caption{Case studies of argumentation-augmented decision-making. (a) Qnt-Arg corrects an IQ2 outcome prediction using the EW-QBAF. (b) Qlt-Arg supplies credibility scores and grounded acceptance annotations with the same five CONFACT passages.}
\label{fig:case-studies}
\end{figure}

\begin{table}[!tbp]
\centering
\caption{Qlt-Arg ablations without GE filtering on CONFACT with Top-10 evidence. $\Delta$ is the reported change in percentage points from the complete configuration.}
\label{tab:confact-ablation-top10}
\small
\begin{tabular*}{\linewidth}{@{\extracolsep{\fill}}ccrr@{}}
\toprule
Credibility scores & Acceptance annotations & Accuracy (\%) & $\Delta$ (pp) \\
\midrule
$\checkmark$ & $\checkmark$ & \textbf{80.33} & $0.00$ \\
--- & $\checkmark$ & 76.72 & $-3.61$ \\
$\checkmark$ & --- & 78.52 & $-1.80$ \\
--- & --- & 78.23 & $-2.10$ \\
\bottomrule
\end{tabular*}
\end{table}

\textbf{Ablation study.}
Table~\ref{tab:confact-ablation-top10} compares the information supplied to the final LLM. Providing both credibility scores and grounded acceptance annotations yields 80.33\% accuracy, whereas the accuracy decreases to 78.52\% with credibility scores alone, 76.72\% with acceptance annotations alone, and 78.23\% with neither. The joint configuration becomes the strongest: numerical credibility for deciding the direction of attacks in AF and relational acceptance information for recording the grounded extension shall be presented together to an LLM for final decision.

\subsection{Case Studies}
\label{sec:case-studies}

Figure~\ref{fig:case-studies} showcases how the two methods correct individual predictions.

\textbf{Qnt-Arg: reasoning about audience change.}
In the IQ2 sample, \textit{For} wins by a larger increase in audience support. The direct decision model assigns the outcome claim 0.3850. Two second-level rebuttals, $b_8^{-}$ and $b_2^{-}$, question the causal support of emotional anecdotes and their ability to convert undecided voters, rather than reinforce existing opposition. They weaken $a_4$ (Richard Pound) and $a_8$ (George Michael), which attack the outcome claim. Setting these two edge weights to zero in the same $D=2$ graph gives $\sigma(r)=0.4918$; restoring their recorded weights gives 0.5261 and the correct prediction \emph{True}.

\textbf{Qlt-Arg: distinguishing causal explanations.}
The CONFACT sample distinguishes direct burning from a tear caused by retching and vomiting. RAG LLM answers \emph{Yes}. CNET headline E4 describes a burn, whereas the same report's body passage E5 and TIME correction E2 attribute the tear to vomiting. Jev scores E5 above E4 (0.62 versus 0.40), yielding that E5 attacks E4, as shown in Figure 4. Grounded semantics accepts E1, E2, and E5 in the full six-edge AF. With all five passages, their credibility scores, and grounded acceptance annotations, the same DeepSeek model correctly answers \emph{No}.

\FloatBarrier
\section{Conclusions}

We abstract a four-stage workflow from existing argumentation-augmented methods and develop two task-specific methods based on it. Qnt-Arg combines speaker-conditioned argumentation construction, separate argument and relation assessments, and candidate-answer priors under gradual semantics. Qlt-Arg uses credibility to orient evidence conflicts and grounded acceptance to guide LLM decision-making. The two argumentation-augmented methods instantiate distinct roles for formal reasoning: computing candidate answer strengths and providing evidence acceptance information. Qnt-Arg reaches 78.41\% overall accuracy across six debate datasets, a 1.42 pp increase compared with 76.99\% for the strongest direct baseline. Qlt-Arg without GE filtering performs best on CONFACT, while GE filtering improves over RAG LLM in all six retrieval settings. The results show the effectiveness of task-specific designs of argumentation construction, model roles, and formal semantic reasoning.

\textbf{Limitations and future work.}
The current experimental coverage remains limited; evaluation on more diverse datasets, domains, and decision tasks is needed to examine generalization. Existing methods, including ours, also use relatively simple forms of integration between LLMs and formal argumentation, and the observed performance gains remain limited. Future work should explore deeper integration, including using semantic reasoning results to revise argumentation construction and assessment, so that formal reasoning can inform multiple stages of the decision process.

%
%
%
%

\end{document}